\documentclass[10pt,twocolumn,letterpaper]{article}

 \usepackage[pagenumbers]{cvpr} % To force page numbers, e.g. for an arXiv version

\usepackage[dvipsnames]{xcolor}

\definecolor{cvprblue}{rgb}{0.21,0.49,0.74}
\usepackage[pagebackref,breaklinks,colorlinks,citecolor=cvprblue]{hyperref}
\usepackage{amsmath}

\def\paperID{*****} % *** Enter the Paper ID here
\def\confName{CVPR}
\def\confYear{2024}

\title{VastGaussian: Vast 3D Gaussians for Large Scene Reconstruction \vspace{-10pt}}
\author{
	Jiaqi Lin\textsuperscript{1$\ast$}\;\;\;
	Zhihao Li\textsuperscript{2$\ast$}\;\;\;
	Xiao Tang\textsuperscript{2}\;\;\;
	Jianzhuang Liu\textsuperscript{3}\;\;\;
	Shiyong Liu\textsuperscript{2}\;\;\;
	Jiayue Liu\textsuperscript{1}\\
	Yangdi Lu\textsuperscript{2}\;\;\;
	Xiaofei Wu\textsuperscript{2}\;\;\;
	Songcen Xu\textsuperscript{2}\;\;\;
	Youliang Yan\textsuperscript{2}\;\;\;
	Wenming Yang\textsuperscript{1\textdagger}\\
	{\textsuperscript{1}Tsinghua University} \;\;\;\; {\textsuperscript{2}Huawei Noah's Ark Lab} \;\;\;\; {\textsuperscript{3}Chinese Academy of Sciences}\\
	{\small $^{\ast}$ Equal contribution \;\;\; $^{\dag}$ Corresponding author}
}

\begin{document}

% \maketitle
\twocolumn[{%
	\renewcommand
	\twocolumn[1][]{#1}%
	\maketitle
	\begin{center}
		\centering
		 \vspace{-15pt}
		\includegraphics[width=\textwidth]{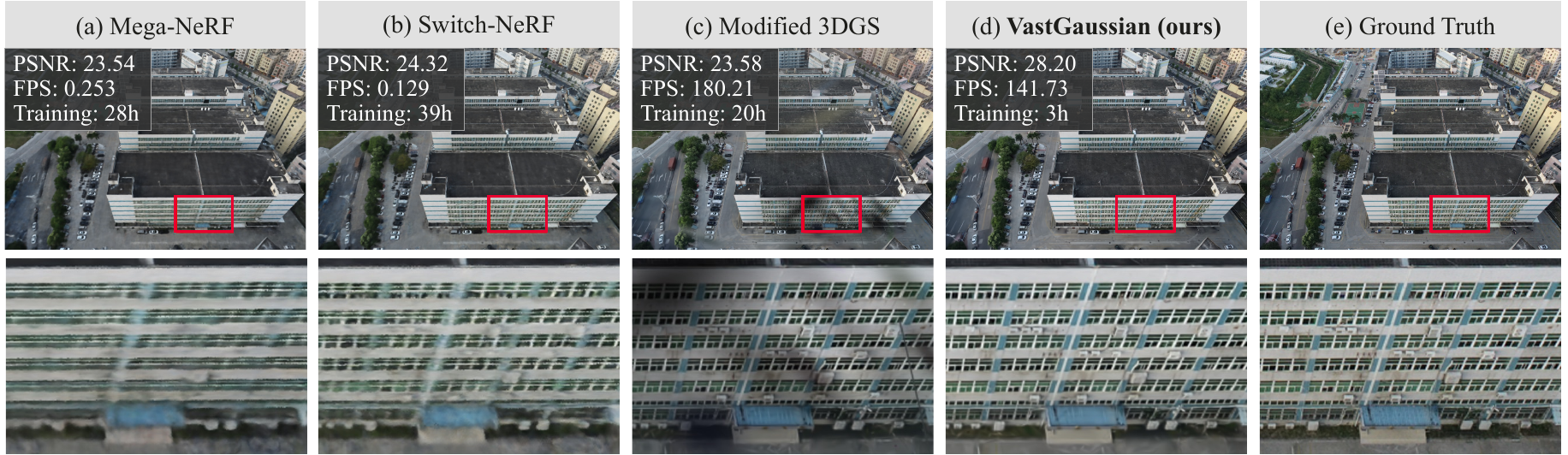}
		\captionof{figure}{Renderings of three state-of-the-art methods and our VastGaussian from the \textit{Residence} scene in the UrbanScene3D dataset \cite{lin2022capturing}. (a, b) Mega-NeRF \cite{turki2022mega} and Switch-NeRF \cite{zhenxing2022switch} produce blurry results with slow rendering speeds. (c) We modify 3D Gaussian Splatting (3DGS) \cite{kerbl20233d} so that it can be optimized for enough iterations on a $32$ GB GPU. The rendered image is much sharper, but with a lot of floaters. (d) Our VastGaussian achieves higher quality and much faster rendering than state-of-the-art methods in large scene reconstruction, with much shorter training time.}
		\label{fig:teaser}
	\end{center}
}]

\begin{abstract}
Existing NeRF-based methods for large scene reconstruction often have limitations in visual quality and rendering speed.
While the recent 3D Gaussian Splatting works well on small-scale and object-centric scenes, scaling it up to large scenes poses challenges due to limited video memory, long optimization time, and noticeable appearance variations.
To address these challenges, we present VastGaussian, the first method for high-quality reconstruction and real-time rendering on large scenes based on 3D Gaussian Splatting.
We propose a progressive partitioning strategy to divide a large scene into multiple cells, where the training cameras and point cloud are properly distributed with an airspace-aware visibility criterion.
These cells are merged into a complete scene after parallel optimization.
We also introduce decoupled appearance modeling into the optimization process to reduce appearance variations in the rendered images.
Our approach outperforms existing NeRF-based methods and achieves state-of-the-art results on multiple large scene datasets, enabling fast optimization and high-fidelity real-time rendering.
Project page: \url{https://vastgaussian.github.io}.
\end{abstract}    
\section{Introduction}
\label{sec:intro}
\begin{figure}[t]
  \centering
   \includegraphics[width=\linewidth]{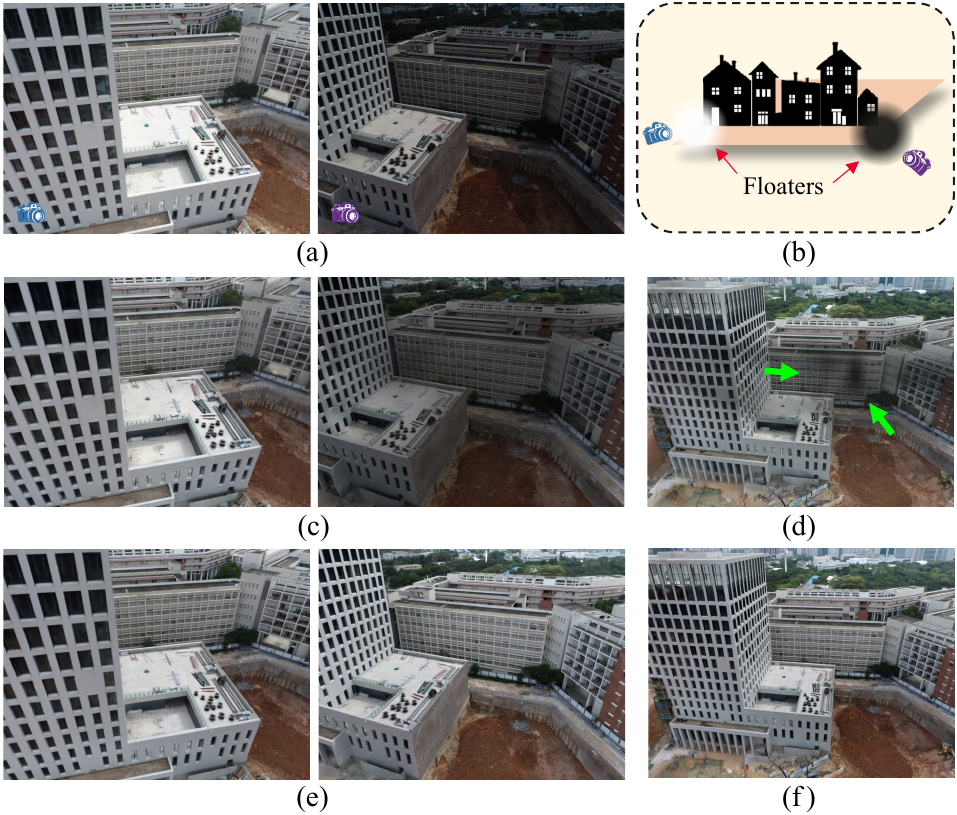}

   \caption{(a) Appearance may vary in adjacent training views. (b)~Dark or bright blobs may be created near cameras with training images of different brightnesses. (c) 3D Gaussian Splatting uses these blobs to fit the appearance variations, making the renderings similar to the training images in (a). (d) These blobs appear as floaters in novel views. (e) Our decoupled appearance modeling enables the model to learn constant colors, so the rendered images are more consistent in appearance across different views. (f) Our approach greatly reduces floaters in novel views.}
   \label{fig:appearance_teaser}
\end{figure}

Large scene reconstruction is essential for many applications, including autonomous driving \cite{li2019aads,yang2020surfelgan,ost2021neural}, aerial surveying \cite{bozcan2020air,du2018unmanned}, and virtual reality, which require photo-realistic visual quality and real-time rendering.
Some approaches \cite{tancik2022block,turki2022mega,xiangli2022bungeenerf,zhenxing2022switch,xu2023grid} are introduced to extend neural radiance fields (NeRF) \cite{mildenhall2021nerf} to large-scale scenes, but they still lack details or render slowly.
Recently, 3D Gaussian Splatting (3DGS) \cite{kerbl20233d} emerges as a promising approach with impressive performance in visual quality and rendering speed, enabling photo-realistic and real-time rendering at 1080p resolution. It is also applied to dynamic scene reconstruction \cite{luiten2023dynamic,yang2023deformable,wu20234d,yang2023real} and 3D content generation \cite{tang2023dreamgaussian,chen2023text,yi2023gaussiandreamer}.
However, these methods focus on small-scale and object-centric scenes.
When applied to large-scale environments, there are several scalability issues.
\textit{First}, the number of 3D Gaussians is limited by a given video memory, while the rich details of a large scene require numerous 3D Gaussians.
Naively applying 3DGS to a large-scale scene would result in either low-quality reconstruction or out-of-memory errors. For intuitive explanation, a $32$ GB GPU can be used to optimize about $11$ million 3D Gaussians, while the small \textit{Garden} scene in the Mip-NeRF 360 dataset \cite{barron2022mip} with an area of less than $100m^2$ already requires approximately $5.8$ million 3D Gaussians for a high-fidelity reconstruction.
\textit{Second}, it requires sufficient iterations to optimize an entire large scene as a whole, which could be time-consuming, and unstable without good regularizations.
\textit{Third}, the illumination is usually uneven in a large scene, and there are noticeable appearance variations in the captured images, as shown in \cref{fig:appearance_teaser}(a).
3DGS tends to produce large 3D Gaussians with low opacities to compensate for these disparities across different views. 
For example, bright blobs tend to come up close to the cameras with images of high exposure, and dark blobs are associated with images of low exposure. These blobs turn to be unpleasant floaters in the air when observed from novel views, as shown in \cref{fig:appearance_teaser}(b, d).

To address these issues, we propose Vast 3D Gaussians (VastGaussian) for large scene reconstruction based on 3D Gaussian Splatting.
We reconstruct a large scene in a divide-and-conquer manner:
Partition a large scene into multiple cells, optimize each cell independently, and finally merge them into a full scene.
It is easier to optimize these cells due to their finer spatial scale and smaller data size. 
A natural and naive partitioning strategy is to distribute training data geographically based on their positions.
This may cause boundary artifacts between two adjacent cells due to few common cameras, and can produce floaters in the air without sufficient supervision.
Thus, we propose visibility-based data selection to incorporate more training cameras and point clouds progressively, which ensures seamless merging and eliminates floaters in the air.
Our approach allows better flexibility and scalability than 3DGS.
Each of these cells contains a smaller number of 3D Gaussians, which reduces the memory requirement and optimization time, especially when optimized in parallel with multiple GPUs.
The total number of 3D Gaussians contained in the merged scene can greatly exceed that of the scene trained as a whole, improving the reconstruction quality.
Besides, we can expand the scene by incorporating new cells or fine-tune a specific region without retraining the entire large scene.

To reduce the floaters caused by appearance variations, Generative Latent Optimization (GLO) \cite{bojanowski2018optimizing} with appearance embeddings \cite{martin2021nerf} is proposed for NeRF-based methods \cite{tancik2022block,zhenxing2022switch}.
This approach samples points through ray-marching, and the point features are fed into an MLP along with appearance embeddings to obtain the final colors.
The rendering process is the same as the optimization, which still requires appearance embeddings as input.
It is not suitable for 3DGS as its rendering is performed by frame-wise rasterization without MLPs.
Therefore, we propose a novel decoupled appearance modeling that is applied only in the optimization.
We attach an appearance embedding to the rendered image pixel-by-pixel, and feed them into a CNN to obtain a transformation map for applying appearance adjustment on the rendered image.
We penalize the structure dissimilarities between the rendered image and its ground truth to learn constant information, while the photometric loss is calculated on the adjusted image to fit the appearance variations in the training image.
Only the consistent rendering is what we need, so this appearance modeling module can be discarded after optimization, thus not slowing down the real-time rendering speed.

Experiments on several large scene benchmarks confirm the superiority of our method over NeRF-based methods.
Our contributions are summarized as follows:
\begin{itemize}
    \item We present VastGaussian, the first method for high-fidelity reconstruction and real-time rendering on large scenes based on 3D Gaussian Splatting. 
    \item We propose a progressive data partitioning strategy that assigns training views and point clouds to different cells, enabling parallel optimization and seamless merging.
    \item We introduce decoupled appearance modeling into the optimization process, which suppresses floaters due to appearance variations. This module can be discarded after optimization to obtain the real-time rendering speed.
\end{itemize}

\section{Related Work}
\label{sec:formatting}
\subsection{Large Scene Reconstruction}
There is significant progress in image-based large scene reconstruction over the past decades.
Some works \cite{agarwal2011building,fruh2004automated,li2008modeling,pollefeys2008detailed,snavely2006photo,zhu2018very,schonberger2016structure} follow a structure-from-motion (SfM) pipeline to estimate camera poses and a sparse point cloud.
The following works \cite{furukawa2010towards,goesele2007multi} produce a dense point cloud or triangle mesh from the SfM output based on multi-view stereo (MVS).
As NeRF \cite{mildenhall2021nerf} becomes a popular 3D representation for photo-realistic novel-view synthesis in recent years \cite{ramamoorthi2023nerfs}, many variants are proposed to improve quality \cite{barron2021mip,barron2022mip,barron2023zip,verbin2022ref,wang2021neus,wang2023neus2,yariv2021volume,li2023neuralangelo,wang2023f2nerf}, increase speed \cite{fridovich2022plenoxels,reiser2023merf,muller2022instant,chen2022tensorf,reiser2021kilonerf,yu2021plenoctrees,sun2022direct,yariv2023bakedsdf,hedman2021baking,chen2023mobilenerf,tang2022nerf2mesh,wang2022r2l,cao2022real}, extend to dynamic scenes \cite{cao2023hexplane,fridovich2023k,liu2023robust,weng_humannerf_2022_cvpr,lin2022efficient,gao2022dynamic}, and so on.
Some methods \cite{tancik2022block,turki2022mega,zhenxing2022switch,xiangli2022bungeenerf,xu2023grid} scale it to large scenes.
Block-NeRF \cite{tancik2022block} divides a city into multiple blocks and distributes training views according to their positions.
Mega-NeRF \cite{turki2022mega} uses grid-based division and assigns each pixel in an image to different grids through which its ray passes.
Unlike these heuristics partitioning strategies, Switch-NeRF \cite{zhenxing2022switch} introduces a mixture-of-NeRF-experts framework to learn scene decomposition.
Grid-NeRF \cite{xu2023grid} does not perform scene decomposition, but rather uses an integration of NeRF-based and grid-based methods.
While the rendering quality of these methods is significantly improved over traditional ones, they still lack details and render slowly.
Recently, 3D Gaussian Splatting \cite{kerbl20233d} introduces an expressive explicit 3D representation with high-quality and real-time rendering at 1080p resolution. However, it is non-trivial to scale it up to large scenes. Our VastGaussian is the first one to do so with novel designs for scene partitioning, optimizing, and merging. 

\subsection{Varying Appearance Modeling}
Appearance variation is a common problem in image-based reconstruction under changing lighting or different camera setting such as auto-exposure, auto-white-balance and tone-mapping.
NRW \cite{meshry2019neural} trains an appearance encoder in a data-driven manner with a contrastive loss, which takes a deferred-shading deep buffer as input and produces an appearance embedding (AE).
NeRF-W \cite{martin2021nerf} attaches AEs to point-based features in ray-marching, and feeds them into an MLP to obtain the final colors, which becomes a standard practice in many NeRF-based methods \cite{tancik2022block,turki2022mega,zhenxing2022switch}. 
Ha-NeRF \cite{chen2022hallucinated} makes AE a global representation across different views, and learns it with a view-consistent loss.
In our VastGaussian, we concatenate AEs with rendered images, feed them into a CNN to obtain transformation maps, and use the transformation maps to adjust the rendered images to fit the appearance variations.

\section{Preliminaries}
\begin{figure*}
	\centering
	\includegraphics[width=\linewidth]{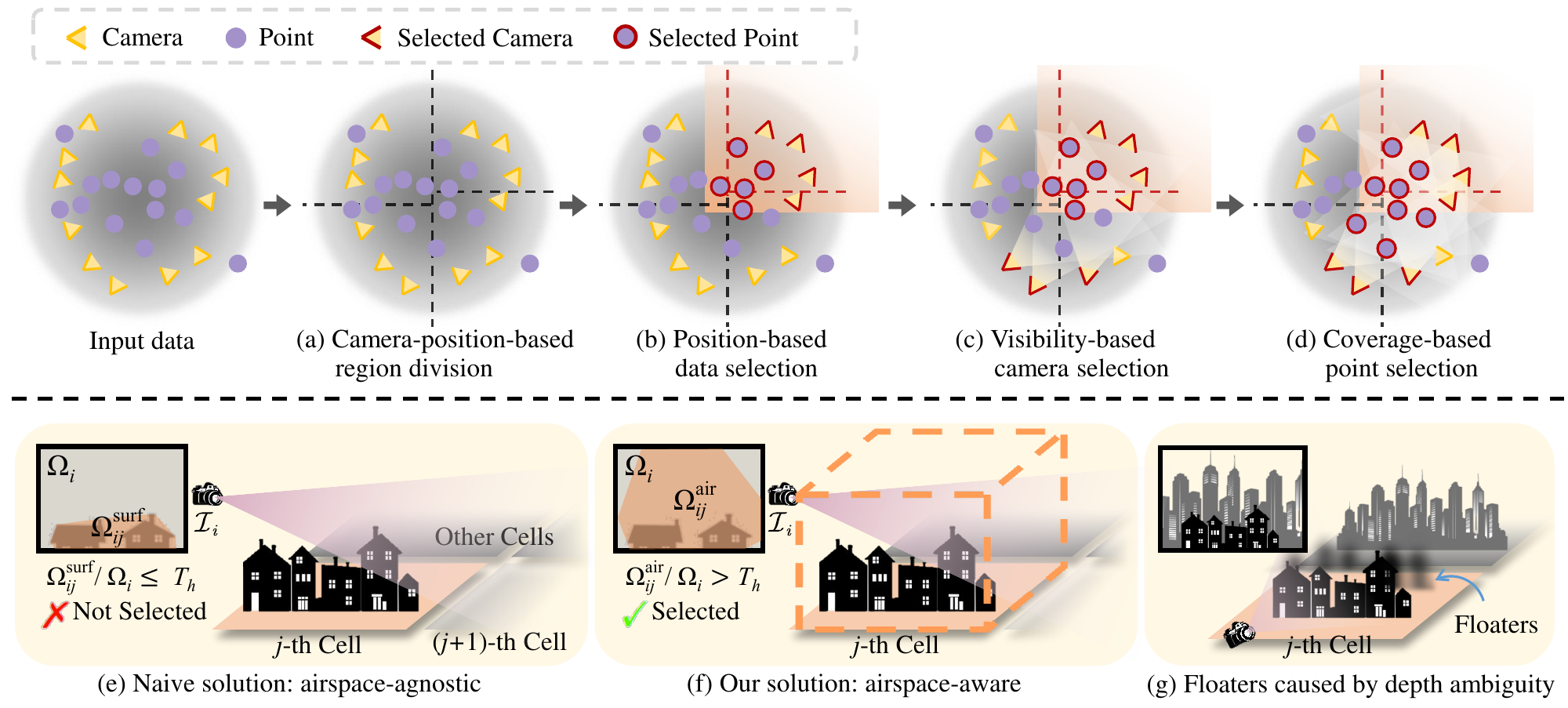}
	\caption{\small \textbf{Progressive data partitioning}. \textbf{Top row}: (a) The whole scene is divided into multiple regions based on the 2D camera positions projected on the ground plane. (b) Parts of the training cameras and point cloud are assigned to a specific region according to its expanded boundaries. (c) More training cameras are selected to reduce floaters, based on an airspace-aware visibility criterion, where a camera is selected if it has sufficient visibility on this region. (d) More points of the point cloud are incorporated for better initialization of 3D Gaussians, if they are observed by the selected cameras. \textbf{Bottom row}: Two visibility definitions to select more training cameras. (e) A naive way: The visibility of the $i$-th camera on the $j$-th cell is defined as $\Omega^\text{surf}_{ij}/\Omega_{i}$, where $\Omega_{i}$ is the area of the image $\mathcal{I}_i$, and $\Omega^\text{surf}_{ij}$ is the convex hull area formed by the surface points in the $j$-th cell that are projected to $\mathcal{I}_i$. (f) Our airspace-aware solution: The convex hull area $\Omega^\text{air}_{ij}$ is calculated on the projection of the $j$-th cell's bounding box in $\mathcal{I}_i$. (g) Floaters caused by depth ambiguity with improper point initialization, which cannot be eliminated without sufficient supervision from training cameras.}
	\label{fig:partition}
\end{figure*}

In this paper, we propose VastGaussian for large scene reconstruction and rendering based on 3D Gaussian Splatting (3DGS) \cite{kerbl20233d}.
3DGS represents the geometry and appearance via a set of 3D Gaussians $\mathbf{G}$.
Each 3D Gaussian is characterized by its position, anisotropic covariance, opacity, and spherical harmonic coefficients for view-dependent colors.
During the rendering process, each 3D Gaussian is projected to the image space as a 2D Gaussian. The projected 2D Gaussians are assigned to different tiles, sorted and alpha-blended into a rendered image in a point-based volume rendering manner \cite{zwicker2001ewa}.

The dataset used to optimize a scene contains a sparse point cloud $\mathbf{P}$ and training views $\mathbf{V}=\left\{ \left( \mathcal{C}_i, \mathcal{I}_i\right)\right\}$, where $\mathcal{C}_i$ is the $i$-th camera, and $\mathcal{I}_i$ is the corresponding image.
$\mathbf{P}$ and $\{\mathcal{C}_i\}$ are estimated by Structure-from-Motion (SfM) from $\{\mathcal{I}_i\}$.
$\mathbf{P}$ is used to initialize 3D Gaussians, and $\mathbf{V}$ is used for differentiable rendering and gradient-based optimization of 3D Gaussians.
For camera $\mathcal{C}_i$, the rendered image $\mathcal{I}^r_i=\mathcal{R}(\mathbf{G}, \mathcal{C}_i)$ is obtained by a differentiable rasterizer $\mathcal{R}$.
The properties of 3D Gaussians are optimized with respect to the loss function between $\mathcal{I}^r_i$ and $\mathcal{I}_i$ as follows:
\begin{equation}
  \mathcal{L}=(1-\lambda)\mathcal{L}_1(\mathcal{I}^r_i, \mathcal{I}_i)+\lambda\mathcal{L}_\text{D-SSIM}(\mathcal{I}^r_i, \mathcal{I}_i),
  \label{eq:loss}
\end{equation}
where $\lambda$ is a hyper-parameter, and $\mathcal{L}_\text{D-SSIM}$ denotes the D-SSIM loss \cite{kerbl20233d}.
This process is interleaved with adaptive point densification, which is triggered when the cumulative gradient of the point reaches a certain threshold.

\section{Method}

3DGS \cite{kerbl20233d} works well on small and object-centric scenes, but it struggles when scaled up to large environments due to video memory limitation, long optimization time, and appearance variations.
In this paper, we extend 3DGS to large scenes for real-time and high-quality rendering.
We propose to partition a large scene into multiple cells that are merged after individual optimization.
In \cref{sec:partition}, we introduce a progressive data partitioning strategy with airspace-aware visibility calculation.
\cref{sec:train} elaborates how to optimize individual cells, presenting our decoupled appearance modeling to capture appearance variations in images.
Finally, we describe how to merge these cells in \cref{sec:merge}.

\subsection{Progressive Data Partitioning}
\label{sec:partition}
We partition a large scene into multiple cells and assign parts of the point cloud $\mathbf{P}$ and views $\mathbf{V}$ to these cells for optimization. 
Each of these cells contains a smaller number of 3D Gaussians, which is more suitable for optimization with lower memory capacity, and requires less training time when optimized in parallel.
The pipeline of our progressive data partitioning strategy is shown in \cref{fig:partition}.

\noindent \textbf{Camera-position-based region division.}
As shown in \cref{fig:partition}(a), we partition the scene based on the projected camera positions on the ground plane, and make each cell contain a similar number of training views to ensure balanced optimization between different cells under the same number of iterations.
Without loss of generality, assuming that a grid of $m\times n$ cells fits the scene in question well, we first partition the ground plane into $m$ sections along one axis, each containing approximately $|\mathbf{V}|/m$ views.
Then each of these sections is further subdivided into $n$ segments along the other axis, each containing approximately $|\mathbf{V}|/(m\times n)$ views.
Although here we take grid-based division as an example, our data partitioning strategy is also applicable to other geography-based division methods, such as sectorization and quadtrees.

\noindent \textbf{Position-based data selection.}
As illustrated in \cref{fig:partition}(b), we assign part of the training views $\mathbf{V}$ and point cloud $\mathbf{P}$ to each cell after expanding its boundaries.
Specifically, let the $j$-th region be bounded in a $\ell^h_j\times \ell^w_j$ rectangle; the original boundaries are expanded by a certain percentage, $20\%$ in this paper, resulting in a larger rectangle of size $(\ell^h_j+0.2\ell^h_j)\times (\ell^w_j+0.2\ell^w_j)$.
We partition the training views $\mathbf{V}$ into $\{\mathbf{V}_j\}^{m\times n}_{j=1}$ based on the expanded boundaries, and segment the point cloud $\mathbf{P}$ into $\{\mathbf{P}_j\}$ in the same way.

\noindent \textbf{Visibility-based camera selection.}
We find that the selected cameras in the previous step
are insufficient for high-fidelity reconstruction,
which can lead to poor detail or floater artifact.
To solve this problem, we propose to add more relevant cameras based on a visibility criterion, as shown in \cref{fig:partition}(c).
Given a yet-to-be-selected camera $\mathcal{C}_i$, let $\Omega_{ij}$ be the projected area of the $j$-th cell in the image $\mathcal{I}_i$, and let $\Omega_{i}$ be the area of $\mathcal{I}_i$;
visibility is defined as $\Omega_{ij}/\Omega_{i}$.
Those cameras with a visibility value greater than a pre-defined threshold $T_h$ are selected.

Note that different ways of calculating $\Omega_{ij}$ result in different camera selections.
As illustrated in \cref{fig:partition}(e), a natural and naive solution is based on the 3D points distributed on the object surface.
They are projected on $\mathcal{I}_i$ to form a convex hull of area $\Omega^\text{surf}_{ij}$.
This calculation is airspace-agnostic because it takes only the surface into account.
Therefore some relevant cameras are not selected due to its low visibility on the $j$-th cell in this calculation, which results in under-supervision for airspace, and cannot suppress floaters in the air.

We introduce an airspace-aware visibility calculation, as shown in \cref{fig:partition}(f).
Specifically, 
an axis-aligned bounding box is formed by the point cloud in the $j$-th cell, whose height is chosen as the distance between the highest point and the ground plane.
We project the bounding box onto $\mathcal{I}_i$ and obtain a convex hull area $\Omega^\text{air}_{ij}$.
This airspace-aware solution takes into account all the visible space, which ensures that given a proper visibility threshold, the views with significant contributions to the optimization of this cell are selected and provide enough supervision for the airspace.

\noindent \textbf{Coverage-based point selection.}
After adding more relevant cameras to the $j$-th cell's camera set $\mathbf{V}_j$, we add the points covered by all the views in $\mathbf{V}_j$ into $\mathbf{P}_j$, as illustrated in \cref{fig:partition}(d).
The newly selected points can provide better initialization for the optimization of this cell.
As illustrated in \cref{fig:partition}(g), some objects outside the $j$-th cell can be captured by some views in $\mathbf{V}_j$, and new 3D Gaussians are generated in wrong positions to fit these objects due to depth ambiguity without proper initialization.
However, by adding these object points for initialization, new 3D Gaussians in correct positions can be easily created to fit these training views, instead of producing floaters in the $j$-th cell.
Note that the 3D Gaussians generated outside the cell are removed after the optimization of the cell.

\subsection{Decoupled Appearance Modeling}
\label{sec:train}
\begin{figure}[t]
	\centering
	\includegraphics[width=0.99\linewidth]{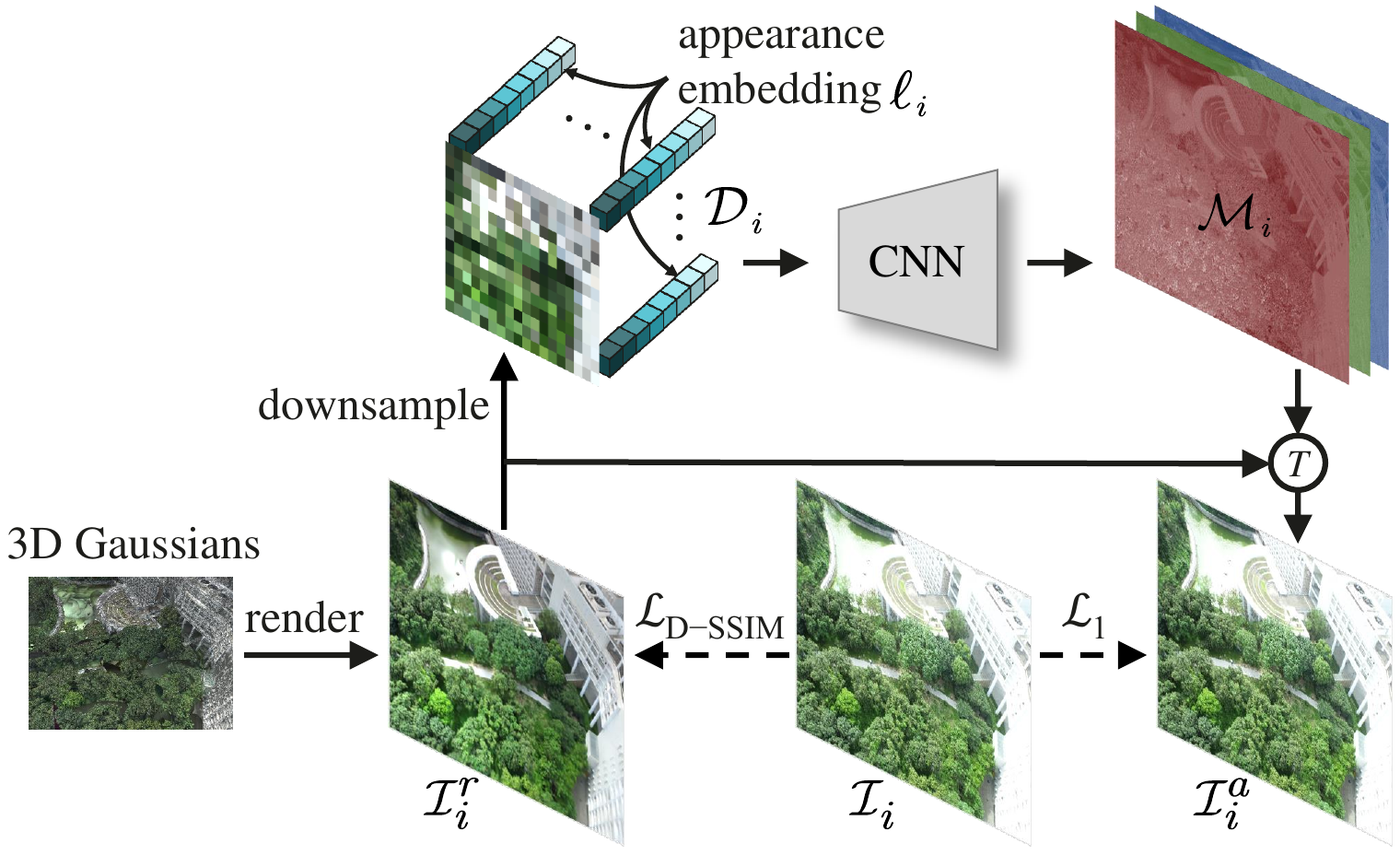}
	
	\caption{Decoupled appearance modeling. The rendered image $\mathcal{I}^r_i$ is downsampled to a smaller resolution, concatenated by an optimizable appearance embedding $\mathbf{\ell }_i$ in a pixel-wise manner to obtain $\mathcal{D}_i$, and then fed into a CNN to generate a transformation map $\mathcal{M}_i$. $\mathcal{M}_i$ is used to perform appearance adjustment on $\mathcal{I}^r_i$ to get an appearance-variant image $\mathcal{I}^a_i$, which is used to calculate the loss $\mathcal{L}_1$ against the ground truth $\mathcal{I}_i$, while $\mathcal{I}^r_i$ is used to calculate the D-SSIM loss.}
	\label{fig:appearance}
\end{figure}
There are obvious appearance variations in the images taken in uneven illumination, and 3DGS tends to produce floaters to compensate for these variations across different views, as shown in \cref{fig:appearance_teaser}(a--d).

To address this problem, some NeRF-based methods \cite{martin2021nerf,tancik2022block,turki2022mega,zhenxing2022switch} concatenate an appearance embedding to point-based features in pixel-wise ray-marching, and feed them into the radiance MLP to obtain the final color.
This is not suitable for 3DGS, whose rendering is performed by frame-wise rasterization without MLPs.
Instead, we introduce decoupled appearance modeling into the optimization process, which produces a transformation map to adjust the rendered image to fit the appearance variations in the training image, as shown in \cref{fig:appearance}.
Specifically, we first downsample the rendered image $\mathcal{I}^{r}_i$ to not only prevent the transformation map from learning high-frequency details, but also reduce computation burden and memory consumption.
We then concatenate an appearance embedding $\mathbf{\ell }_i$ of length $m$ to every pixel in the three-channel downsampled image, and obtain a 2D map $\mathcal{D}_i$ with $3+m$ channels.
$\mathcal{D}_i$ is fed into a convolutional neural network (CNN), which progressively upsamples $\mathcal{D}_i$ to generate $\mathcal{M}_i$ that is of the same resolution as $\mathcal{I}^r_i$.
Finally, the appearance-variant image $\mathcal{I}^a_i$ is obtained by performing a pixel-wise transformation $T$ on $\mathcal{I}^r_i$ with $\mathcal{M}_i$:
\begin{equation}
	\mathcal{I}^a_i=T(\mathcal{I}^r_i; \mathcal{M}_i).
	\label{eq:affine}
\end{equation}
In our experiments, a simple pixel-wise multiplication works well on the datasets we use.
The appearance embeddings and CNN are optimized along with the 3D Gaussians, using the loss function modified from \cref{eq:loss}:
\begin{equation}
	\mathcal{L}=(1-\lambda)\mathcal{L}_1(\mathcal{I}^a_i, \mathcal{I}_i)+\lambda\mathcal{L}_\text{D-SSIM}(\mathcal{I}^r_i, \mathcal{I}_i).
	\label{eq:loss2}
\end{equation}

Since $\mathcal{L}_\text{D-SSIM}$ mainly penalizes the structural dissimilarity, applying it between $\mathcal{I}^r_i$ and the ground truth $\mathcal{I}_i$ makes the structure information in $\mathcal{I}^r_i$ close to $\mathcal{I}_i$, leaving the appearance information to be learned by $\ell_i$ and the CNN.
The loss $\mathcal{L}_1$ is applied between the appearance-variant rendering $\mathcal{I}^a_i$ and $\mathcal{I}_i$, which is used to fit the ground truth image $\mathcal{I}_i$ that may have appearance variations from other images.
After training, $\mathcal{I}^r_i$ is expected to have a consistent appearance with other images, from which the 3D Gaussians can learn an average appearance and correct geometry of all the input views.
This appearance modeling can be discarded after optimization, without slowing down the real-time rendering speed.

\subsection{Seamless Merging}
\label{sec:merge}
After optimizing all the cells independently, we need to merge them to get a complete scene. 
For each optimized cell, we delete the 3D Gaussians that are outside the original region (\cref{fig:partition}(a)) before boundary expansion.
Otherwise, they could become floaters in other cells.
We then merge the 3D Gaussians of these non-overlapping cells.
The merged scene is seamless in appearance and geometry without obvious border artifacts, because some training views are common between adjacent cells in our data partitioning.
Therefore, there is no need to perform further appearance adjustment like Block-NeRF \cite{tancik2022block}.
The total number of 3D Gaussians contained in the merged scene can greatly exceed that of the scene trained as a whole, thus improving the reconstruction quality.

\begin{table*}[t!]
	\begin{center}
		\resizebox{\linewidth}{!}{
			\begin{tabular}{l|ccc|ccc|ccc|ccc|ccc}
				\toprule
				Scene   &   \multicolumn{3}{c|}{\emph{Building}}  &   \multicolumn{3}{c|}{\emph{Rubble}} &   \multicolumn{3}{c|}{\emph{Campus}} &  \multicolumn{3}{c|}{\emph{Residence}} &   \multicolumn{3}{c}{\emph{Sci-Art}} \\
				\midrule
				Metrics &  SSIM & PSNR & LPIPS   &
				SSIM & PSNR & LPIPS   &
				SSIM & PSNR & LPIPS   &
				SSIM & PSNR & LPIPS   &
				SSIM & PSNR & LPIPS     \\
				\midrule
				Mega-NeRF & 0.569 & 21.48 & 0.378 & 0.575  & 24.70 & 0.407 & 0.561 & 23.93 & 0.513 & 0.648 & 22.86 & 0.330 & 0.769 & 26.25 & 0.263 \\
				Switch-NeRF  & 0.594 & 22.07 & 0.332 & 0.586 & 24.93  & 0.377 & 0.565 & 24.03 & 0.495 & 0.675 & 23.41 & 0.280 & 0.793 & \tb{27.07} & 0.224\\
				Grid-NeRF (grid branch) & -- & -- & -- & 0.780  & 25.47 & 0.213 & \underline{0.767} & \underline{25.51} & 0.174 & 0.807 & \tb{24.37} & 0.142 & -- & -- & --\\
				Grid-NeRF (nerf branch) & -- & -- & -- & 0.767  & 24.13 & 0.207 & 0.757 & 24.90 & \underline{0.162} & 0.802 & 23.77 & \underline{0.137} & -- & -- & -- \\
				Modified 3DGS  & \underline{0.769}  & \underline{23.01} & \underline{0.164} & \underline{0.800}  & \underline{26.78} & \underline{0.161} & 0.712 & 23.89 & 0.289 & \underline{0.825} & 23.40 & 0.142 & \underline{0.843} & 25.24 & \underline{0.166}\\
				\midrule
				\tb{VastGaussian (Ours)} & \tb{0.804} & \tb{23.50} & \tb{0.130} & \tb{0.823} & \tb{26.92} & \tb{0.132} & \tb{0.816} & \tb{26.00} & \tb{0.151} & \tb{0.852} & \underline{24.25} & \tb{0.124} & \tb{0.885} & \underline{26.81} & \tb{0.121}\\
				\bottomrule
			\end{tabular}		
		}
		\caption{Quantitative evaluation of our method compared to previous work on five large scenes. We report SSIM$\uparrow$, PSNR$\uparrow$ and LPIPS$\downarrow$ on test views. The \tb{best} and \underline{second best} results are highlighted. ``--" denotes missing data from the Grid-NeRF paper.}
		\label{tab:compare}
		\vspace{-3mm}
	\end{center}
	\centering
\end{table*}
\section{Experiments}
\subsection{Experimental Setup}
\noindent \textbf{Implementation.}
We evaluate our model with $8$ cells in our main experiments.
The visibility threshold is $25\%$.
The rendered images are downsampled by $32$ times before being concatenated with the appearance embeddings of length $64$.
Each cell is optimized for $60,000$ iterations.
The densification \cite{kerbl20233d} starts at the $1,000$th iteration and ends at the $30,000$th iteration, with an interval of $200$ iterations.
The other settings are identical to those of 3DGS \cite{kerbl20233d}.
Both the appearance embeddings and the CNN use a learning rate of $0.001$.
We perform Manhattan world alignment to make the $y$-axis of the world coordinate perpendicular to the ground plane.
We describe the CNN architecture in the supplement.

\noindent \textbf{Datasets.}
The experiments are conducted on five large-scale scenes: \textit{Rubble} and \textit{Building} from the Mill-19 dataset \cite{turki2022mega}, and \textit{Campus}, \textit{Residence}, and \textit{Sci-Art} from the UrbanScene3D dataset \cite{lin2022capturing}.
Each scene contains thousands of high-resolution images.
We downsample the images by $4$ times for training and validation, following previous methods \cite{turki2022mega,zhenxing2022switch} for fair comparison.

\noindent \textbf{Metrics.}
We evaluate the rendering quality using three quantitative metrics: SSIM, PSNR, and AlexNet-based LPIPS.
The aforementioned photometric variation makes the evaluation difficult, as it is uncertain which photometric condition should be replicated.
To address this issue, we follow Mip-NeRF 360 \cite{barron2022mip} to perform color correction on the rendered images before evaluating the metrics of all the methods, which solves a per-image least squares problem to align the RGB values between the rendered image and its corresponding ground truth.
We also report the rendering speed at 1080p resolution, average training time, and video memory consumption.

\noindent \textbf{Compared methods.}
We compare our VastGaussian with four methods: Mega-NeRF \cite{turki2022mega}, Switch-NeRF \cite{zhenxing2022switch}, Grid-NeRF \cite{xu2023grid}, and 3DGS \cite{kerbl20233d}.
For 3DGS, we need to increase optimization iterations to make it comparable in our main experiments, but naively doing that causes out-of-memory errors.
Therefore, we increase the densification interval accordingly to build a feasible baseline (termed Modified 3DGS).
The other configurations are the same as those in the original 3DGS paper.
For Grid-NeRF, its code is released without rendered images and carefully tuned configuration files due to its confidentiality requirements.
These unavailable files are critical to its performance, making its results not reproducible.
Therefore, we use its code to evaluate only its training time, memory and rendering speed, while the quality metrics are copied from its paper.

\subsection{Result Analysis}
\begin{figure*}[t!]
	\centering
	\includegraphics[width=\linewidth]{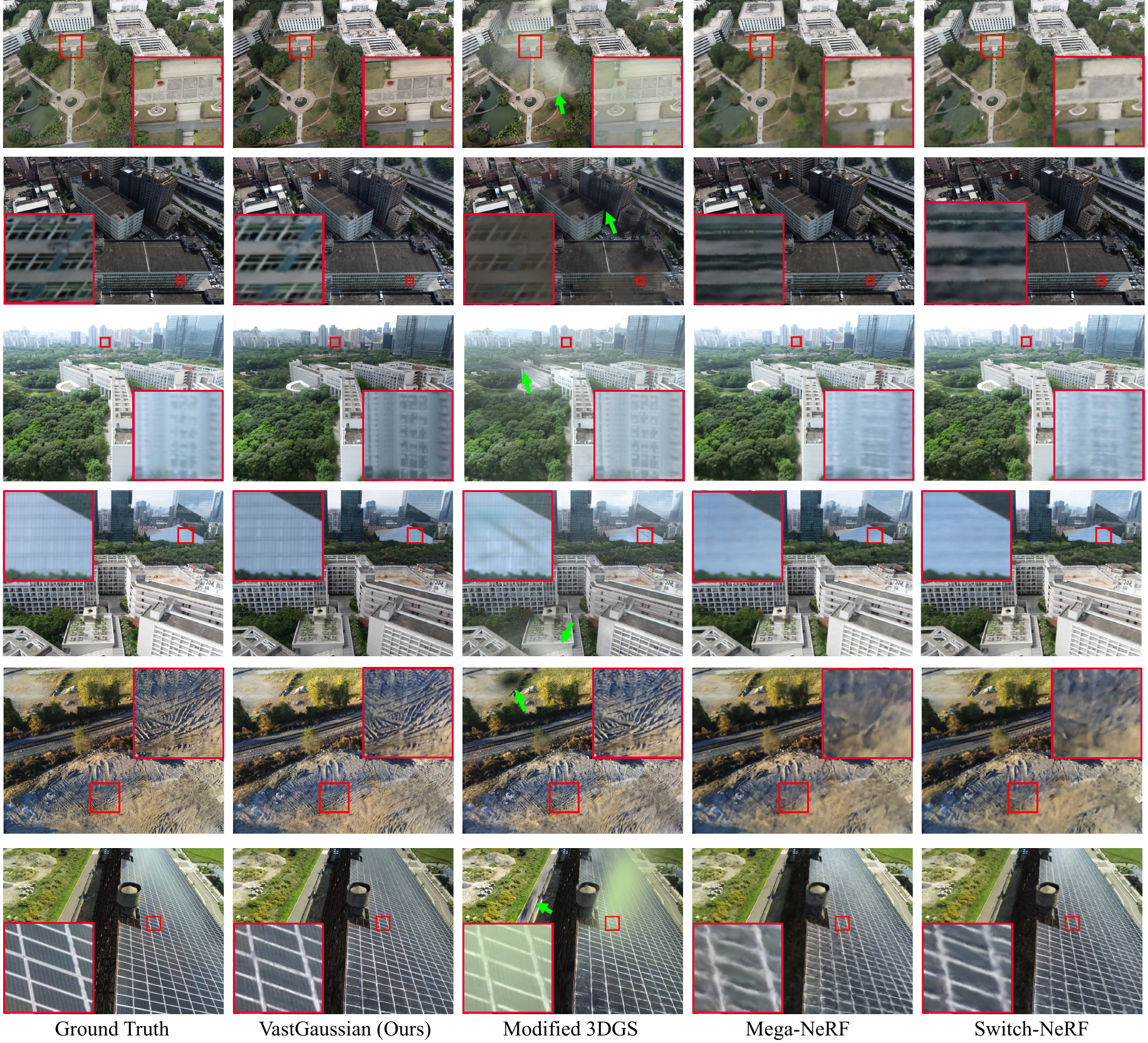}
	\caption{Qualitative comparison between VastGaussian and previous work. Floaters are pointed out by green arrows.}
	\label{fig:compare}
\end{figure*}
\begin{table}[t!]
	\begin{center}
		\resizebox{\linewidth}{!}{
			\begin{tabular}{l|ccc|ccc}
				\toprule
				Dataset   &   \multicolumn{3}{c|}{Mill-19}  &   \multicolumn{3}{c}{UrbanScene3D} \\
				\midrule
				Metrics & Training & VRAM & FPS &  Training & VRAM  & FPS \\
				\midrule
				Mega-NeRF   & 30h19m & \tb{5.0G} & 0.25 & 28h01m & \tb{5.0G} & 0.31   \\
				Switch-NeRF & 41h58m & 9.9G & 0.12 & 38h47m & 9.9G & 0.15\\
				Grid-NeRF   & 17h44m & 14.0G & 0.26 & 17h38m & 14.0G & 0.30 \\
				Modified 3DGS  & 19h32m & 31.1G & \tb{177.75} & 20h12m & 31.2G & \tb{210.99} \\
				\midrule
				\tb{VastGaussian}  & \tb{2h25m} & 10.4G & 126.45 & \tb{2h56m} & 11.9G & 171.92 \\
				\bottomrule
			\end{tabular}		
		}
		\caption{Comparison of training time, training video memory consumption (VRAM), and rendering speed.}
		\label{tab:time}
	\end{center}
	\centering
\end{table}
\textbf{Reconstruction quality.} In \cref{tab:compare}, we report the mean SSIM, PSNR, and LPIPS metrics in each scene.
Our VastGaussian outperforms the compared methods in all the SSIM and LPIPS metrics by significant margins, suggesting that it reconstructs richer details with better rendering in perception. 
In terms of PSNR, VastGaussian achieves better or comparable results.
We also show visual comparison in \cref{fig:compare}.
The NeRF-based methods fall short of details and produce blurry results. Modified 3DGS has sharper rendering but produces unpleasant floaters. Our method achieves clean and visually pleasing rendering.
Note that due to the noticeable over-exposure or under-exposure in some test images, VastGaussian exhibits slightly lower PSNR values, but produces significantly better visual quality, sometimes even being more clear than the ground truth, such as the example on the $3$rd row in \cref{fig:compare}.
The high quality of VastGaussian is partly thanks to its large number of 3D Gaussians. Take the \textit{Campus} scene for example, the number of 3D Gaussians in Modified 3DGS is $8.9$ million, while for VastGaussian the number is $27.4$ million.

\noindent\textbf{Efficiency and memory.}
In \cref{tab:time}, we report the training time, video memory consumption during optimization, and rendering speed.
Mega-NeRF, Switch-NeRF and VastGaussian are trained on 8 Tesla V100 GPUs, while Grid-NeRF and Modified 3DGS on a single V100 GPU as they do not perform scene decomposition.
The rendering speeds are tested on a single RTX 3090 GPU.
Our VastGaussian requires much shorter time to reconstruct a scene with photo-realistic rendering.
Compared to Modified 3DGS, VastGaussian greatly reduces video memory consumption on a single GPU.
Since VastGaussian has more 3D Gaussians in the merged scene than Modified 3DGS, its rendering speed is slightly slower than Modified 3DGS, but is still much faster than the NeRF-based methods, achieving real-time rendering at 1080p resolution.

\subsection{Ablation Study}
We perform ablation study on the \textit{Sci-Art} scene to evaluate different aspects of VastGaussian.

\begin{figure}[t!]
	\centering
	\includegraphics[width=\linewidth]{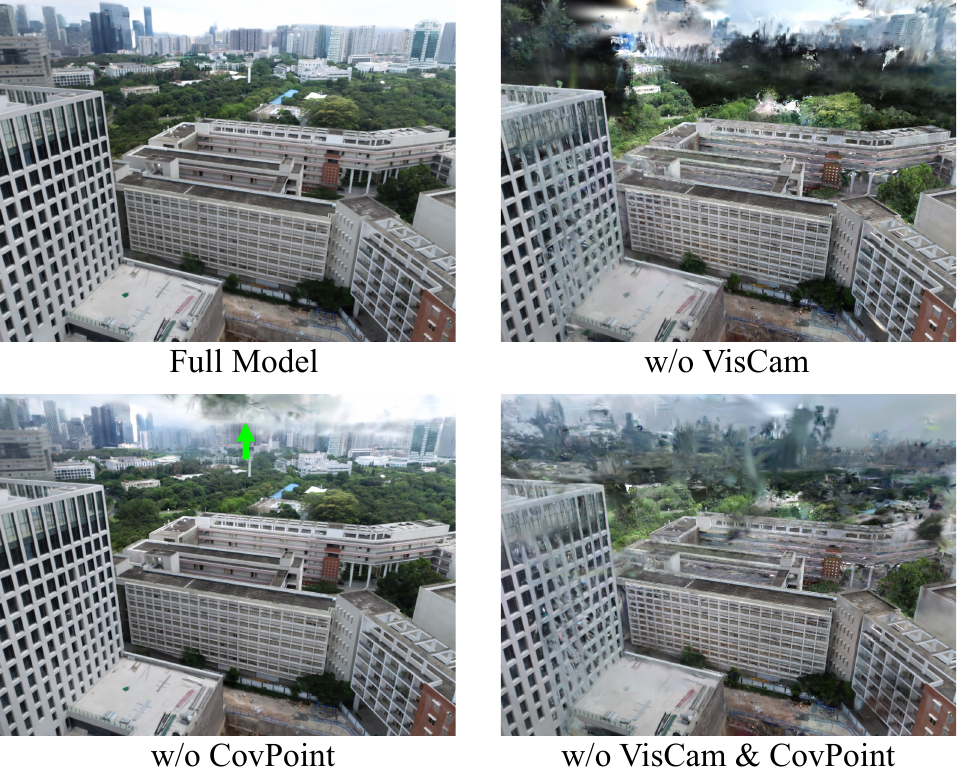}
	
	\caption{The visibility-based camera selection and coverage-based point selection can reduce floaters in the airspace.}
	\label{fig:partition_ablation}
\end{figure}

\begin{figure}[t!]
	\centering
	\includegraphics[width=\linewidth]{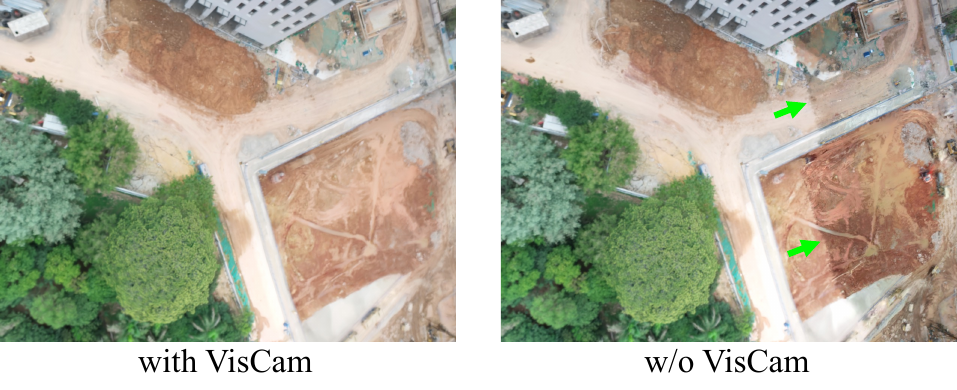}
	
	\caption{The visibility-based camera selection can eliminate the appearance jumping on the cell boundaries.}
	\label{fig:boundary}
\end{figure}

\noindent \textbf{Data partition.} As shown in \cref{fig:partition_ablation} and \cref{tab:partition_ablation}, both visibility-based camera selection (VisCam) and coverage-based point selection (CovPoint) can improve visual quality.
Without each or both of them, floaters can be created in the airspace of a cell to fit the views observing regions outside the cell.
As shown in \cref{fig:boundary}, the visibility-based camera selection can ensure more common cameras between adjacent cells, which eliminates the noticeable boundary artifact of appearance jumping when it is not implemented.

\begin{table}[t!]
    \begin{center}
	    \resizebox{0.9\linewidth}{!}{
		        \begin{tabular}{l|ccc}
			            \toprule
			            Model setting &  SSIM & PSNR & LPIPS \\
			            \midrule
			            1)\ \ w/o VisCam & 0.694 & 20.05 & 0.261\\
			            2)\ \ w/o CovPoint & 0.874 & 26.14 & 0.128\\
			            3)\ \ w/o VisCam \& CovPoint & 0.699 & 20.35 & 0.253\\
			            4)\ \ airspace-aware $\rightarrow$ agnostic & 0.855 & 24.54 & 0.128\\
			            5)\ \ w/o Decoupled AM & 0.858 & 25.08 & 0.148\\
			            \midrule
			            Full model & \tb{0.885} & \tb{26.81} & \tb{0.121} \\
			            \bottomrule
			        \end{tabular}		
		    }
	    \caption{Ablation on data partition, visibility calculation and decoupled appearance modeling (Decoupled AM).}
	    \label{tab:partition_ablation}
	    \vspace{-3mm}
	    \end{center}
\centering
\end{table}

\begin{figure}[t!]
	\centering
	\includegraphics[width=\linewidth]{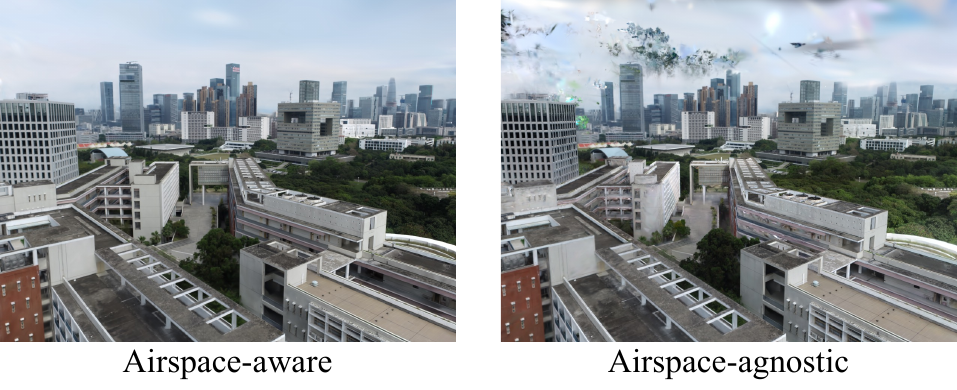}
	
	\caption{Heavy floaters appear when the visibility is calculated in the airspace-agnostic way.}
	\label{fig:airspace_ablation}
\end{figure}

\noindent \textbf{Airspace-aware visibility calculation.} As illustrated in the $4$th row of \cref{tab:partition_ablation} and \cref{fig:airspace_ablation}, the cameras selected based on the airspace-aware visibility calculation provide more supervision for the optimization of a cell, thus not producing floaters that are presented when the visibility is calculated in the airspace-agnostic way.

\noindent\textbf{Decoupled appearance modeling.} As shown in \cref{fig:appearance_teaser} and the $5$th row of \cref{tab:partition_ablation}, our decoupled appearance modeling reduces the appearance variations in the rendered images.
Therefore, the 3D Gaussians can learn consistent geometry and colors from training images with appearance variations, instead of creating floaters to compensate for these variations.
Please also see the videos in the supplement.


\noindent \textbf{Different number of cells}.
As shown in \cref{tab:cell_number}, more cells reconstruct better details in VastGaussian, leading to better SSIM and LPIPS values, and shorter training time when the cells are optimized in parallel.
However, when the cell number reaches $16$ or bigger, the quality improvement becomes marginal, and PSNR slightly decreases because there may be gradual brightness changes in a rendered image from cells that are far apart.

\begin{table}[t!]
	\begin{center}
		\resizebox{0.8\linewidth}{!}{
			\begin{tabular}{c|c|cccc}
				\toprule
				\#Cell & \#GPU &  SSIM & PSNR & LPIPS & Training \\
				\midrule
				4 & 4 & 0.870 & 26.39 & 0.136 &  2h46m \\
				8 & 8 & 0.885 & \tb{26.81} & 0.121 &  2h39m \\
				16 & 16 & 0.888 & 26.80 & 0.116 &  2h30m \\
				24 & 24 & \tb{0.892} & 26.64 & \tb{0.110} &  \tb{2h19m} \\
				\bottomrule
			\end{tabular}		
		}
		\caption{Effect of different cell numbers.}
		\label{tab:cell_number}
		\vspace{-3mm}
	\end{center}
	\centering
\end{table}

\section{Conclusion and Limitation}
In this paper, we propose VastGaussian, the first high-quality reconstruction and real-time rendering method on large-scale scenes. 
The introduced progressive data partitioning strategy allows for independent cell optimization and seamless merging, obtaining a complete scene with sufficient 3D Gaussians.
Our decoupled appearance modeling decouples appearance variations in the training images, and enables consistent rendering across different views. 
This module can be discarded after optimization to obtain faster rendering speed.
While our VastGaussian can be applied to spatial divisions of any shape, we do not provide an optimal division solution that should consider the scene layout, the cell number and the training camera distribution.
In addition, there are a lot of 3D Gaussians when the scene is huge, which may need a large storage space and significantly slow down the rendering speed.
\newpage
{
    \small
    \bibliographystyle{ieeenat_fullname}
    \bibliography{main}
}
% WARNING: do not forget to delete the supplementary pages from your submission 
 \clearpage
\maketitlesupplementary

We elaborate more details of decoupled appearance modeling, including the CNN architecture, training time and memory, exploration of more complex transformations, and visualization of the transformation maps.

\section{Details of Decoupled Appearance Modeling}
\subsection{CNN Architecture}
\begin{figure}[t!]
	\centering
	\includegraphics[width=0.6\linewidth]{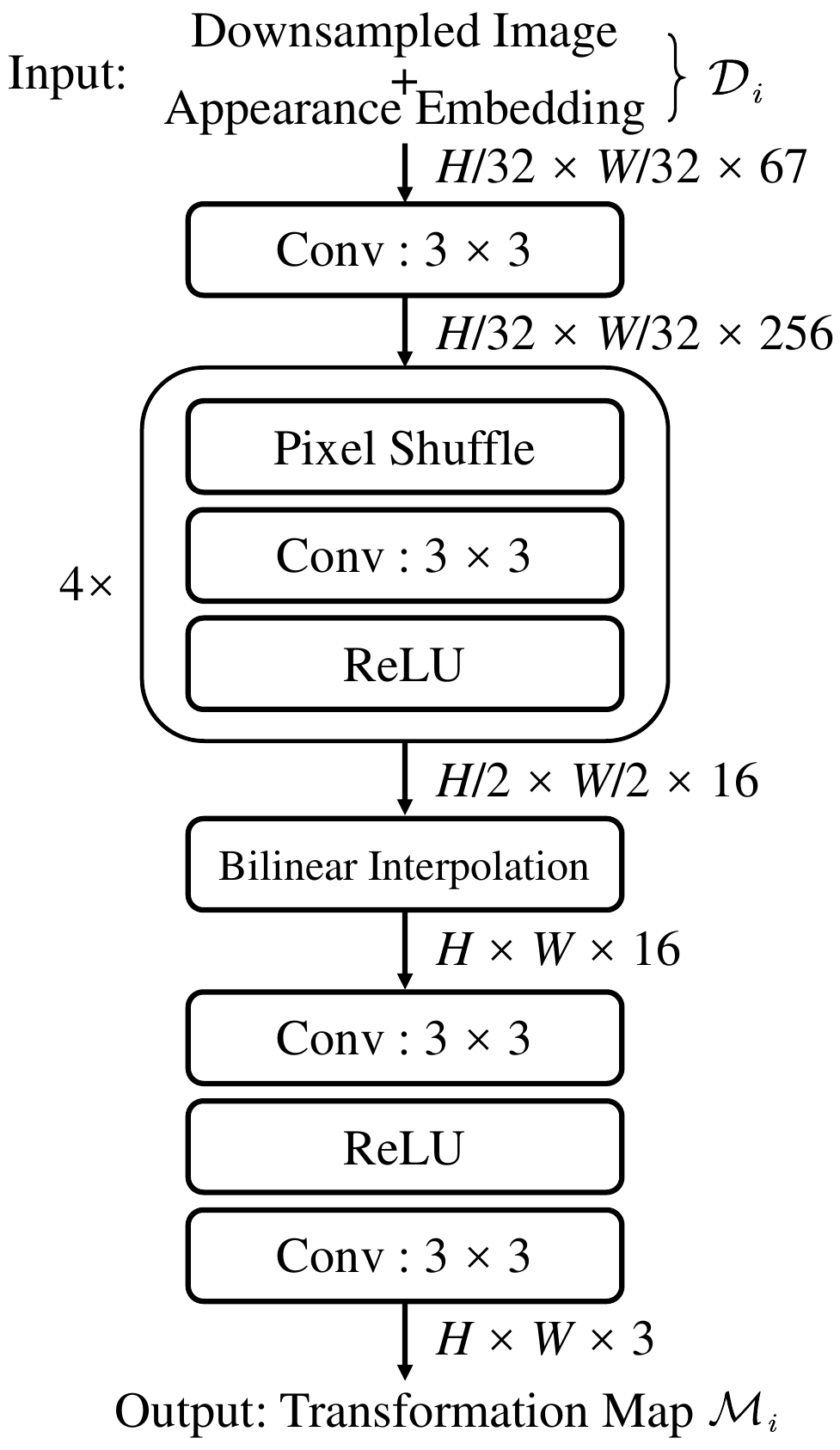}
	
	\caption{CNN architecture of decoupled appearance modeling.}
	\label{fig:cnn}
\end{figure}

The CNN architecture in the decoupled appearance modeling is shown in \cref{fig:cnn}. 
We downsample the rendered image of shape $H\times W \times 3$ by $32$ times, and then concatenate an appearance embedding of length $64$ to every pixel in the downsampled image to obtain a feature map $\mathcal{D}_i$ of shape $\frac{H}{32}\times \frac{W}{32} \times 67$, which is fed to the CNN.

$\mathcal{D}_i$ is first passed through a $3\times 3$ convolutional layer to increase its channel depth to $256$.
Then it is processed by $4$ upsampling blocks, each containing an upsampling pixel shuffle layer, a $3\times 3$ convolutional layer, and $\rm{ReLU}$ activation.
Each upsampling block doubles the feature map resolution and halves its channel depth.
After that, a bilinear interpolation layer upsamples the feature map to $H\times W$, and two $3\times 3$ convolutional layers with a $\rm{ReLU}$ in between are followed to obtain a transformation map of shape $H\times W \times 3$, which is later used to perform appearance adjustment on the rendered image.

\subsection{Training Time and Memory}
We provide the training time and video memory consumption to complement the ablation study of the decoupled appearance modeling in \cref{tab:appearance_time}. Our model with the decoupled appearance modeling takes less than $10\%$ more time and video memory to optimize due to the introduction of the appearance embeddings and the CNN, but it significantly improves rendering quality.

\subsection{More Complex Transformations}
A simple pixel-wise multiplication works well on the datasets we use.
As shown in \cref{tab:transform_ablation}, we can also extend it to the affine transformation or add some prior knowledge such as Gamma correction. 
The CNN architecture is accordingly adjusted to output a $6$-channel map (for the multiplication plus addition) or a $4$-channel map (with one additional channel for Gamma correction).
Both extensions have marginal effect on the results, which indicate that the simple pixel-wise multiplication is sufficient for the  datasets used in the experiments.
\begin{table}[t]
	\begin{center}
		\resizebox{\linewidth}{!}{
			\begin{tabular}{l|ccccc}
				\toprule
				Model setting & SSIM & PSNR & LPIPS & Training & VRAM \\
				\midrule
				w/o Decoupled AM & 0.858 & 25.08 & 0.148 &\tb{1h25m} & \tb{10.23G}\\
				\midrule
				w/ Decoupled AM & \tb{0.885} & \tb{26.81} & \tb{0.121} & 1h33m & 11.18G \\
				\bottomrule
			\end{tabular}		
		}
		\caption{Quality metrics, training time, and video memory in the ablation study of decoupled appearance modeling on the \textit{Sci-Art} scene.}
		\label{tab:appearance_time}
		\vspace{-6mm}
	\end{center}
	\centering
\end{table}
\begin{table}[t]
	\begin{center}
		\resizebox{\linewidth}{!}{
			\begin{tabular}{l|ccc}
				\toprule
				Transformation & SSIM & PSNR & LPIPS \\
				\midrule
				multiplication \quad & 0.885 & 26.81 & 0.116\\
				multiplication + addition & 0.886 & 26.80 & 0.116\\
				multiplication + Gamma correction & 0.885 & 26.83 & 0.115\\
				\bottomrule
			\end{tabular}		
		}
		\caption{More complex transformations of decoupled appearance modeling on the \textit{Sci-Art} scene.}
		\label{tab:transform_ablation}
		\vspace{-6mm}
	\end{center}
	\centering
\end{table}
\begin{figure*}[t]
	\centering
	\includegraphics[width=\linewidth]{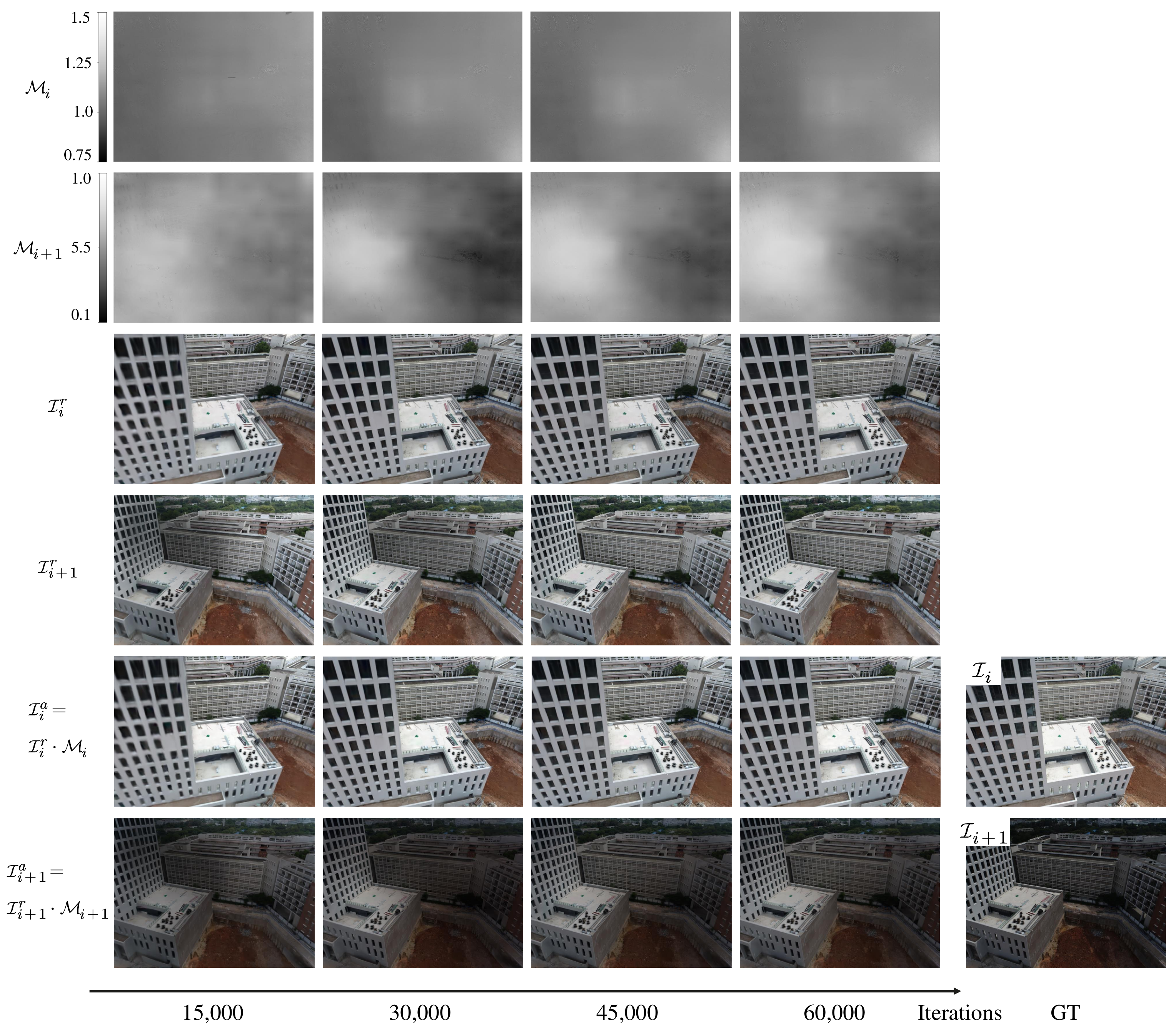}
	\caption{Evolution of the transformation maps ($\mathcal{M}_{i}$ and $\mathcal{M}_{i+1}$), the rendered images ($\mathcal{I}^r_{i}$ and $\mathcal{I}^r_{i+1}$), and the adjusted images ($\mathcal{I}^a_{i}$ and $\mathcal{I}^a_{i+1}$) during the optimization process.}
	\label{fig:appearance_vis}
\end{figure*}
\subsection{Transformation Visualization}
In the main paper, we show that our decoupled appearance modeling enables VastGaussian to learn constant colors between adjacent training views.
For two adjacent training images $\mathcal{I}_{i}$ and $\mathcal{I}_{i+1}$, here we visualize their transformation maps $\mathcal{M}_{i}$ and $\mathcal{M}_{i+1}$, rendered images $\mathcal{I}^r_{i}$ and $\mathcal{I}^r_{i+1}$, and adjusted images $\mathcal{I}^a_{i}$ and $\mathcal{I}^a_{i+1}$, as shown in \cref{fig:appearance_vis}.
As the optimization proceeds, the appearance variations between $\mathcal{I}^r_{i}$ and $\mathcal{I}^r_{i+1}$ are reduced, the appearance transformation maps $\mathcal{M}_{i}$ and $\mathcal{M}_{i+1}$ gradually converge, and the adjusted images $\mathcal{I}^a_{i}$ and $\mathcal{I}^a_{i+1}$ finally fit the training images $\mathcal{I}_{i}$ and $\mathcal{I}_{i+1}$.
Note that for simplicity, we turn the $3$-channel transformation maps into grayscale images in the visualization.

\end{document}